\documentclass[12pt, final]{l4dc2022}

\title[Neighborhood Mixup Experience Replay]{Neighborhood Mixup Experience Replay: \\ Local Convex Interpolation for Improved Sample Efficiency \\in Continuous Control Tasks}
\usepackage{times}

\usepackage{multirow}

\usepackage[style=authoryear, backend=biber]{biblatex}
\usepackage{mleftright}

\usepackage{graphics}
\usepackage{graphicx}

\usepackage{amsmath} 
\usepackage{amssymb}  

\usepackage{algorithm}
\usepackage{algorithmic}

\usepackage[font=small,skip=6pt]{caption}

\author{%
 \Name{Ryan Sander$^1$} \Email{rmsander@mit.edu}\\
 \Name{Wilko Schwarting$^1$} \Email{wilkos@mit.edu}\\
 \Name{Tim Seyde$^1$} \Email{tseyde@mit.edu}\\
 \Name{Igor Gilitschenski$^{2, 3}$} \Email{gilitschenski@cs.toronto.edu}\\
 \Name{Sertac Karaman$^{4}$} \Email{sertac@mit.edu}\\
 \Name{Daniela Rus$^{1}$} \Email{rus@mit.edu}\\
 \addr $^1$CSAIL - MIT, $^2$University of Toronto, $^3$Toyota Research Institute, $^4$LIDS - MIT
}

\begin{document}
\maketitle

\begin{abstract}
Experience replay plays a crucial role in improving the sample efficiency of deep reinforcement learning agents. Recent advances in experience replay propose using Mixup \parencite{DBLP:conf/iclr/ZhangCDL18} to further improve sample efficiency via synthetic sample generation. We build upon this technique with Neighborhood Mixup Experience Replay (NMER), a geometrically-grounded replay buffer that interpolates transitions with their closest neighbors in state-action space. NMER preserves a locally linear approximation of the transition manifold by only applying Mixup between transitions with vicinal state-action features. Under NMER, a given transition’s set of state-action neighbors is dynamic and episode agnostic, in turn encouraging greater policy generalizability via inter-episode interpolation. We combine our approach with recent off-policy deep reinforcement learning algorithms and evaluate on continuous control environments. We observe that NMER improves sample efficiency by an average 94\% (TD3) and 29\% (SAC) over baseline replay buffers, enabling agents to effectively recombine previous experiences and learn from limited data.
\end{abstract}

\section{Introduction}
Learning robust and effective behavior from a limited set of examples is a hallmark of human cognition \parencite{fong2018using}. The sample efficiency of our neuronal circuits allows us to quickly learn new skills even with limited experience. In many problem domains, human sample efficiency far outperforms that of deep reinforcement learning algorithms \parencite{lee2019sample}. Narrowing this gap is a critical milestone toward replicating intelligence in reinforcement learning agents.

Model-free (MF), off-policy deep reinforcement learning (DRL) agents provide significant sample efficiency gains relative to their on-policy counterparts \parencite{gu2017deep,arulkumaran2017brief}. Improved sample efficiency is due largely to experience replay techniques \parencite{mnih2013playing}, which enable agents to learn from past experience. The trial-and-error nature of reinforcement learning nonetheless necessitates collecting large volumes of training data \parencite{pmlr-v100-yang20a, ijcai2018-820,48707}. While lower sample efficiency may be acceptable for learning in simulation, it may significantly hinder an agent's progress in many real-world applications where samples are expensive to generate \parencite{pmlr-v100-yang20a}. 

Model-based (MB) DRL agents achieve improved sample efficiency by learning a model of the environment \parencite{weyand2016planet,DBLP:conf/nips/HaS18,Hafner2020Dream}
that can be used for offline planning and policy refinement~\parencite{browne2012survey, schwarting2021deep, pmlr-v120-seyde20a}.
%
For reinforcement learning tasks with noisy and high-dimensional state and action spaces, however, an agent's learned environment models may suffer from estimation bias when little data is available. Combined with model capacity limitations, this can result in model-based agents converging to suboptimal policies  \parencite{48707,RENAUDO2015178}.

We aim to combine the benefits of learning on true environment interactions in MF-DRL with the sample efficiency benefits of MB-DRL. To this end, we propose Neighborhood Mixup Experience Replay (NMER), a modular replay buffer that improves the sample efficiency of off-policy, MF-DRL agents by training on experiences sampled from convex, linear combinations of vicinal transitions from the replay buffer. NMER interpolates neighboring pairs of transitions in the geometric transition space of the replay buffer using Mixup \parencite{DBLP:conf/iclr/ZhangCDL18}, a convex and stochastic linear interpolation technique. Despite the computational and analytical simplicity of Mixup, its use in experience replay can improve generalization and policy convergence through implicit regularization and expansion of the training support of the agent's neural network function approximators. We empirically observe these benefits of Mixup in our continuous control experiments.

NMER can be applied to any continuous control reinforcement learning agent leveraging experience replay. As a motivating example, consider a robotic humanoid learning to walk using off-policy, MF-DRL agents, given a limited set of experiences consisting of odometry and actuator sensor measurements. With standard experience replay \parencite{mnih2013playing} approaches, the finite size of the agent's experiences can hinder the agent from learning robust policies, perhaps due to the agent not experiencing a crucial subset of the transition space. With NMER, however, interpolated experiences can provide this DRL agent with crucial training samples in regions of the transition space they previously did not experience, thus improving the robustness and performance of the policies the agent learns. Our contributions\footnote{Code for NMER can be found at \href{https://github.com/rmsander/interreplay}{this GitHub repository.}} are thus summarized as follows:
\begin{enumerate}
\label{para:contributions}
    \item \textbf{Neighborhood Mixup Experience Replay (NMER)}: A geometrically-grounded replay buffer that improves the sample efficiency of off-policy, MF-DRL agents by training these agents on linear combinations of vicinal transitions.
    \item \textbf{Local Mixup}: A generalization of NMER, this algorithm considers the application of Mixup between vicinal points in any feature space, with proximity defined by a distance metric.
    \item \textbf{Improved sample efficiency in continuous control}: Our evaluation study demonstrates that NMER substantially improves sample efficiency of off-policy, MF-DRL algorithms across several continuous control environments.
\end{enumerate}

\section{Related work}
\label{sec:related_work}
Experience replay, data augmentation, and interpolation approaches have been applied to RL and other machine learning domains. NMER builds off of these techniques to improve sample efficiency.

\paragraph{Experience replay} \textit{Prioritized Experience Replay} (PER) \parencite{schaul2016prioritized} samples an agent's experiences from a replay buffer according to the ``learnability'' or ``surprise'' that each sample induces in the agent in its current parameterization. PER uses absolute TD-error of a sample \parencite{schaul2016prioritized} as a heuristic measure of ``surprise''. In the stochastic prioritization variant of PER, transitions are sampled proportionally to their learnability. While this technique improves the sample efficiency by selecting highly-relevant samples, it does not improve the overall ``learnability'' of the samples themselves and restricts training to previously observed experience. \textit{Experience Replay Optimization} (ERO) \parencite{ijcai2019-589} parameterizes the replay buffer directly as a learned priority score function. Rather than using heuristics such as TD-error to determine a prioritization of samples, as is performed in PER \parencite{schaul2016prioritized}, in ERO this prioritization is learned directly via a policy gradient approach in which return is measured by the agent's policy improvement \parencite{ijcai2019-589}. A REINFORCE-based \parencite{williams1992simple} estimate of the policy gradient updates the learned replay buffer in an alternating fashion with the policy being trained. Similarly to PER, while ERO improves sample efficiency by selecting samples with high ``learnability'', it does not improve the overall ``learnability'' of the samples themselves, and also restricts training to the agent's observed experiences. \textit{Interpolated Rewards Replay} \parencite{von2020bootstrapping} performs linear interpolation of experienced rewards. In contrast, NMER interpolates entire transitions using stochastic convex linear interpolation, resulting in a more expressive interpolation of an agent's experiences. 

\paragraph{Data augmentation} Data augmentation techniques are also used to improve the performance of DRL agents. \textit{Reinforcement learning with Augmented Data (RAD)} is a module designed for improving agent performance in visual and propioceptive DRL tasks \parencite{NEURIPS2020_e615c82a}. For continuous control environments, RAD leverages techniques such as random amplitude scaling (RAS). While RAS does allow for learning beyond an agent's observed set of experiences, it does not consider meaningful combinations of these experiences. In \textit{Data-regularized Q (DrQ)} learning, geometric-invariant data augmentation mechanisms are applied to off-policy DRL algorithms to improve sample efficiency in visual control tasks, providing off-policy agents with sample efficiency comparable to state-of-the-art MB-DRL algorithms \parencite{kostrikov2020image}. Similar to DrQ, NMER improves sample efficiency via regularization through training agents on augmented samples.

\paragraph{Mixup sampling} Mixup was originally applied to supervised machine learning domains, and empirically improves the generalizability and out-of-sample predictive performance of learners \parencite{DBLP:conf/iclr/ZhangCDL18}. Several reinforcement learning methodologies make use of Mixup-interpolated experiences for training reinforcement learning agents. In \textit{Continuous Transition} \parencite{lin2020continuous}, temporally-adjacent transitions are interpolated with Mixup, generating synthetic transitions between pairs of consecutive transitions. In \textit{MixReg} \parencite{wang2020improving}, generated transitions are formed using Mixup on combinations of input and output signals. In \textit{S4RL} \parencite{sinha2021srl}, generated transitions are produced by interpolating current ($s_t$) and next ($s_{t+1}$) states within an observed transition. While these approaches increase the training domain via interpolation, they do not strictly enforce geometric transition proximity of the resulting samples. Proximity between the points used for sampling is encoded temporally, as in \parencite{lin2020continuous, sinha2021srl}, but not in the geometric transition space of the agent's experience. NMER employs a nearest neighbor heuristic to encourage transition pairs for Mixup to be located approximately within the same dynamics regimes in the transition manifold. Compared to Continuous Transition \parencite{lin2020continuous} and S4RL \parencite{sinha2021srl}, samples interpolated with NMER may better preserve the local dynamics of the environment and enable further agent regularization through inter-episode interpolation between transitions and their dynamic sets of nearest neighbors.

\section{Preliminaries}
Neighborhood Mixup Experience Replay (NMER) builds on experience replay for off-policy DRL, Mixup, and nearest neighbor heuristics to encourage approximately on-manifold interpolation.
%


\paragraph{Off-policy DRL for continuous control tasks}
Off-policy DRL has successfully been applied to continuous control tasks through the use of actor-critic methods such as Soft Actor-Critic (SAC), Deep Deterministic Policy Gradients (DDPG), and Twin Delayed DDPG (TD3) \parencite{pmlr-v80-haarnoja18b, journals/corr/LillicrapHPHETS15, fujimoto2018addressing}. In this off-policy, MF-DRL setting, agents are trained using transitions composed of states, actions, rewards, and next states. The state ($\mathcal{S}$), action ($\mathcal{A}$), and reward ($\mathcal{R}$) spaces that define these transitions are continuous.

\paragraph{Experience replay.} Experience replay \parencite{mnih2013playing} enables an agent to train on past observations. It can be largely decoupled from the agent's training algorithm - while the agent seeks to learn optimal policies and value functions given observed training samples, regardless of the samples provided to it, the experience replay buffer is tasked with providing the agent samples that offer the greatest ``learnability'' for improving these policies and value functions. Current experience replay approaches are discussed in Section \ref{sec:related_work}.

\paragraph{Mixup.} Mixup \parencite{DBLP:conf/iclr/ZhangCDL18} is a novel stochastic data augmentation technique that improves the generalizability of supervised learners by training them on convex linear combinations of existing samples. This linear interpolation mechanism invokes a prior on the learner that linear combinations of features result in the same linear combinations of targets \parencite{DBLP:conf/iclr/ZhangCDL18}, leading to generalizability in accordance with Occam's Razor \parencite{rasmussen2001occam}. Through another lens, Mixup increases the generalizability of a learner by increasing its training support with interpolated transitions. These transitions are sampled from the convex unit line connecting two transitions \parencite{DBLP:conf/iclr/ZhangCDL18} according to a symmetric beta distribution parameterized by $\alpha$, which controls the spread of the distribution along this relative unit line. To interpolate a new sample $\mathbf{x}_{\text{interpolated}}$ using two existing samples $\mathbf{x}_1, \mathbf{x}_2 \in \mathbb{R}^{d}$, Mixup interpolates according to:
\begin{equation}
    \mathbf{x}_{\text{interpolated}} = \lambda\mathbf{x}_1 + (1-\lambda)\mathbf{x}_2, \; \lambda \sim \beta(\alpha, \alpha), \; \alpha > 0
\end{equation}

\begin{figure}[ht]
    \centering
    \includegraphics[width=0.4\textwidth]{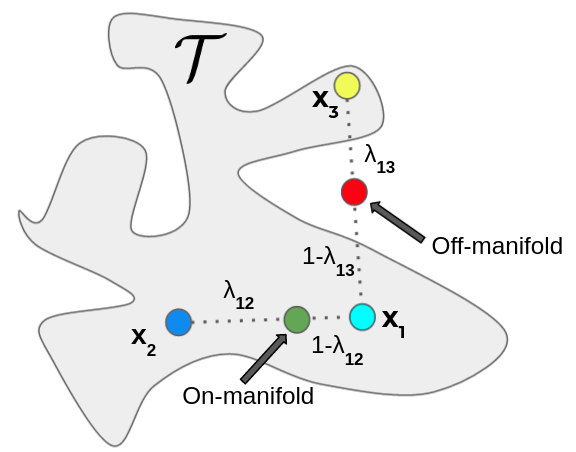}
    \caption{Examples of on and off-transition manifold interpolation using Mixup. On-manifold or approximately on-manifold interpolation is crucial for successfully training DRL agents in continuous control tasks.}
    \label{fig:on_manifold}
\end{figure}

\paragraph{On-manifold interpolation.} To measure the accuracy of interpolation in interpolated experience replay approaches, we consider how ``on-manifold'' the interpolated transition is with respect to the transition manifold mapping states and actions to rewards and next states. We consider that the observed transitions of a replay buffer lie on a transition manifold, which we denote by the space $\mathcal{T}$. The space $\mathcal{T}$ is given by the Cartesian product of the state ($\mathcal{S}$), action ($\mathcal{A}$), reward ($\mathcal{R}$), and next state ($\mathcal{S}$) spaces of the agent as: $\mathcal{T}: \mathcal{S} \times \mathcal{A} \times \mathcal{R} \times \mathcal{S}$. With NMER, the use of Mixup between only proximal/neighboring transitions acts as a heuristic to encourage interpolated transitions to remain near the underlying transition manifold, as depicted by Figure \ref{fig:on_manifold}. See our \href{https://rmsander.github.io/projects/nmer_tech_report.pdf}{technical report} for details on on-manifold assessment.
\section{Neighborhood Mixup Experience Replay (NMER)}
NMER trains off-policy MF-DRL agents using convex linear combinations of an agent's existing, proximal experiences, effectively creating locally linear models centered around each transition of the replay buffer. By only interpolating proximal transitions with one another, where proximity is measured by the standardized Euclidean distance in the state-action space of the replay buffer, NMER interpolates transitions that have similar state and action inputs, but potentially different reward and next state outputs. In considering these nearest neighbors, NMER regularizes the off-policy MF-DRL agents it trains by allowing inter-episode interpolation between proximal transitions. Furthermore, in the presence of stochasticity in the transition manifold, NMER can prevent these agents from overfitting to a particular (reward, next state) outcome by interpolating different (reward, next state) outcomes for near-identical (state, action) inputs. NMER consists of two steps:
\begin{enumerate}
    \item \textbf{Update step}: When a new environment interaction is added to the replay buffer, re-standardize the states and actions of the stored transitions in the replay buffer, and update the nearest neighbor data structures using Euclidean distances over the Z-score standardized, concatenated state-action features of the replay buffer. Similarity search is thus measured over the input state and action spaces; however, we emphasize that NMER can admit other distance functions and representations of similarity as well. See our \href{https://rmsander.github.io/projects/nmer_tech_report.pdf}{technical report} for further details.
    \item \textbf{Sampling Step}: First, we sample a batch of ``sample transitions'' uniformly from the replay buffer. Next, we query the nearest neighbors of each transition in this sampled batch. Following this, for each set of neighbors in the training batch, we sample a neighbor transition uniformly from this set of neighbors, and apply Mixup to linearly interpolate each pair of selected samples and neighbors ($\mathbf{x}_{\text{sample}, i}$ and $\mathbf{x}_{\text{neighbor}, i}$, respectively):
        \begin{equation}
        \mathbf{x}_{\text{interpolated}, i} = \lambda\mathbf{x}_{\text{sample}, i} + (1-\lambda)\mathbf{x}_{\text{neighbor, i}}, \; \lambda \sim \beta(\alpha, \alpha), \alpha > 0
    \end{equation}
\end{enumerate}

\begin{figure}[ht]
    \centering
    \includegraphics[width=0.75\textwidth]{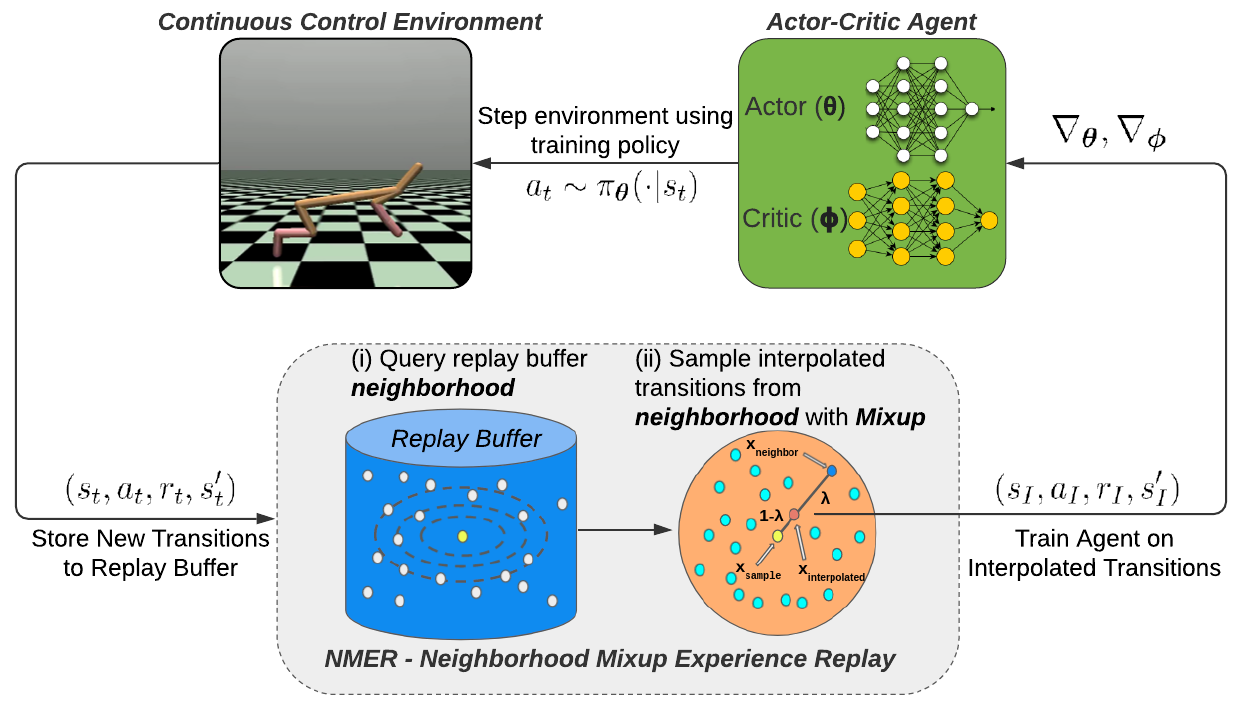}
    \caption{With NMER, convex interpolation is performed between a sampled transition and its neighboring transitions, improving the generalizability and robustness of off-policy, MF-DRL agents applied to continuous control tasks. }
    \label{fig:sys_nmer}
\end{figure}

These steps are given in Algorithm \ref{alg:nmer_unbatched}, and depicted in Figure \ref{fig:sys_nmer}. NMER introduces minimal computational overhead compared to standard experience replay, requiring only vectorized standardization, nearest neighbor querying, and local Mixup operations. This positions NMER as a viable experience replay buffer for high-dimensional continuous control tasks.
\begin{algorithm}
\textbf{Input}: Replay buffer $\mathcal{B}$, Mixup hyperparameter $\alpha > 0$, Batch Size T \\
\textbf{Output}: Interpolated training batch $\textbf{B}_{\text{train}}$ \\
\caption{Neighborhood Mixup Experience Replay (NMER)}
\begin{algorithmic}
\STATE $\mathbf{B} = \{(s_t, a_t, r_t, s'_t)\}_{t=1}^{\text{T}} \overset{\text{iid}}{\sim} \mathcal{U}(\mathcal{B})$ \;\;\COMMENT{\;\text{Sample Batch uniformly from Replay Buffer}}
\STATE $\mathbf{B}_{\text{train}} \leftarrow \text{Array}[\cdots]$
\FOR{t in T}
\STATE $(s_s, a_s, r_s, s'_s) \leftarrow \mathbf{B}[t]$ \;\;\COMMENT{\;\text{Sampled Transition for NMER}}
\STATE $\mleft[\tilde{s}_s, \tilde{a}_s \mright]^T \leftarrow \textbf{ZScore}(\mleft[s_s, a_s \mright]^T)$ \;\;\COMMENT{\;\text{Standardize States and Actions of Sampled Transition}}
\STATE $K_{s} \leftarrow$ \textbf{NN}$\left(\mleft[\tilde{s}_s, \tilde{a}_s \mright]^T, \mathcal{B}\right)$ \;\;\COMMENT{\;\text{Standardized Local Neighborhood of Sampled Transition}}
\STATE $(s_n, a_n, r_n, s'_n) \sim \mathcal{U}(K_{s})$ \;\;\COMMENT{\;\text{Sample Neighboring Transition from Local Neighborhood}}
\STATE $\lambda \sim \beta(\alpha, \alpha)$ \;\;\COMMENT{\;\text{Sample Mixup Coefficient}}

\STATE $\mathbf{x}_s \leftarrow \mleft[s_s, a_s, r_s, s'_s \mright]^T$  \;\;\COMMENT{\; Sampled Transition Features}
\STATE $\mathbf{x}_n \leftarrow \mleft[s_n, a_n, r_n, s'_n \mright]^T$  \;\;\COMMENT{\; Neighboring Transition Features}
\STATE $\mathbf{x}_i = \lambda \mathbf{x}_s + (1-\lambda)\mathbf{x}_n$ \;\;\COMMENT{\;\text{Interpolate Sampled and Neighboring Transitions using Mixup}}
\STATE $\mathbf{B}_{\text{train}}[t] \leftarrow \mathbf{x}_i$ \;\;\COMMENT{\;\text{Add Interpolated Sample to Training Batch}}
\ENDFOR
\RETURN $\mathbf{B}_{\text{train}}$
\end{algorithmic}
\label{alg:nmer_unbatched}
\end{algorithm}

\paragraph{Agent regularization via linear interpolation.} Through the lens of Occam's Razor \parencite{rasmussen2001occam}, NMER improves the generalizability of the policy and value function approximators of off-policy, MF-DRL agents by invoking the prior that linear combinations of state-action pairs result in the same linear combinations of corresponding reward-next state pairs. This prior improves generalizability in tasks where this linearity assumption approximately holds. Since the spaces of an agent in continuous control tasks are continuous, interpolating continuous, linear combinations of transitions can still yield interpolated samples that lie proximate to the underlying transition manifold $\mathcal{T}$.

Furthermore, if the transition manifold $\mathcal{T}$ is convex, NMER guarantees on-manifold interpolation, since this technique generates strictly convex combinations of transitions. In this regime, synthetically-generated on-manifold transitions are indistinguishable from transitions generated at the same point using the underlying environment dynamics. However, for many applications, particularly high-dimensional, real-world continuous control tasks, the underlying transition manifold will generally be non-convex.

\paragraph{Neighborhood Mixup as a heuristic to encourage on-manifold interpolation.} Non-convexity and nonlinearity in continuous control environments provide motivation for our neighborhood-based interpolation mechanism, which addresses issues with non-convexity of the transition manifold by only considering interpolation between transitions in the same ``neighborhood'', i.e. transitions with similar state-action pairs. If the transition manifold is locally Euclidean, linearly interpolating two transitions is a suitable, approximately on-manifold mechanism for interpolating between spatially proximal transitions.

\section{Continuous control evaluation}
To rigorously evaluate and quantify the improvement in sample efficiency with NMER, we compare NMER to other state-of-the-art replay buffers by applying these replay buffers to continuous control tasks and off-policy, MF-DRL algorithms.

\textbf{Testing environment and configuration.} \label{para:eval_process} We consider continuous control environments from the OpenAI Gym MuJoCo \parencite{1606.0154,todorov2012mujoco} suite. Since we are principally interested in evaluating the sample efficiency of replay buffers, we treat the off-policy, MF-DRL algorithms used for these evaluations (SAC \parencite{pmlr-v80-haarnoja18b} and TD3 \parencite{fujimoto2018addressing}) as part of the experimental configuration. For each replay buffer variant, including NMER, we train agents using replay or update-to-data ratios (ratio of gradient steps to environment interactions) of 1, 5, and 20, and report the best results for each replay buffer variant. Additionally, for SAC \parencite{pmlr-v80-haarnoja18b}, to stabilize the policy, we add a small L2 regularizer to the actor network for all SAC evaluations. Implementation details and ablation studies for each replay buffer variant are provided in the \href{https://rmsander.github.io/projects/nmer_tech_report.pdf}{technical report}. We measure replay buffer sample efficiency using the evaluation reward of the reinforcement learning agent after 200K environment interactions have been sampled, as in \parencite{lin2020continuous, lee2020sunrise}. Rewards are smoothed using an averaging window of 11, as in \parencite{SpinningUp2018}. 

\textbf{Baselines.} \label{sec:baseline} We compare NMER to the following baselines: (i) \textit{Uniform, Vanilla Replay} (\texttt{U}) \parencite{mnih2013playing,10.1145/1102351.1102377}, where transitions are sampled i.i.d. uniformly from the replay buffer, (ii) \textit{Prioritized Experience Replay} (\texttt{PER}) \parencite{schaul2016prioritized} with stochastic prioritization, (iii) \textit{Continuous Transition} (\texttt{CT}) \parencite{lin2020continuous}. Since the main comparison between NMER and CT is how samples are selected for interpolation, we make two modifications to the original CT baseline: (a) We remove the automatic Mixup $\alpha$ hyperparameter tuning mechanism, and (b) If a terminal state is encountered in either the sample or neighbor transition, no interpolation occurs, and the sampled transition is simply used for training the agent. In Continuous Transition, terminal transitions in an episode can be interpolated with their previous, non-terminal transitions, resulting in non-binary termination signals. In order to make the fairest comparison possible between NMER and CT, we adopt the same interpolation rules we use for NMER in CT. (iv) \textit{SUNRISE Baselines}. Additionally, we compare the performance of NMER to SUNRISE baselines, tested using the same continuous control environments and 200K environment interaction evaluation \parencite{lee2020sunrise}. (v) \textit{Neighborhood Size Limits of NMER} (\texttt{1NN-Mixup} and \texttt{Mixup}, respectively), which represent the limits of NMER with one neighbor ($k=1$) and all neighbors ($k=|\mathcal{B}|$), respectively. (vi) \textit{S4RL} (\texttt{S4RL}), which implements the replay buffer interpolation technique from \parencite{sinha2021srl}. Lastly, (vii) \textit{Noisy Replay} ($\mathcal{N}(0, \sigma^2)$), which trains DRL agents on transitions with i.i.d. Gaussian noise added to each component. 

\begin{figure}[ht]
    \centering
    \includegraphics[width=\textwidth]{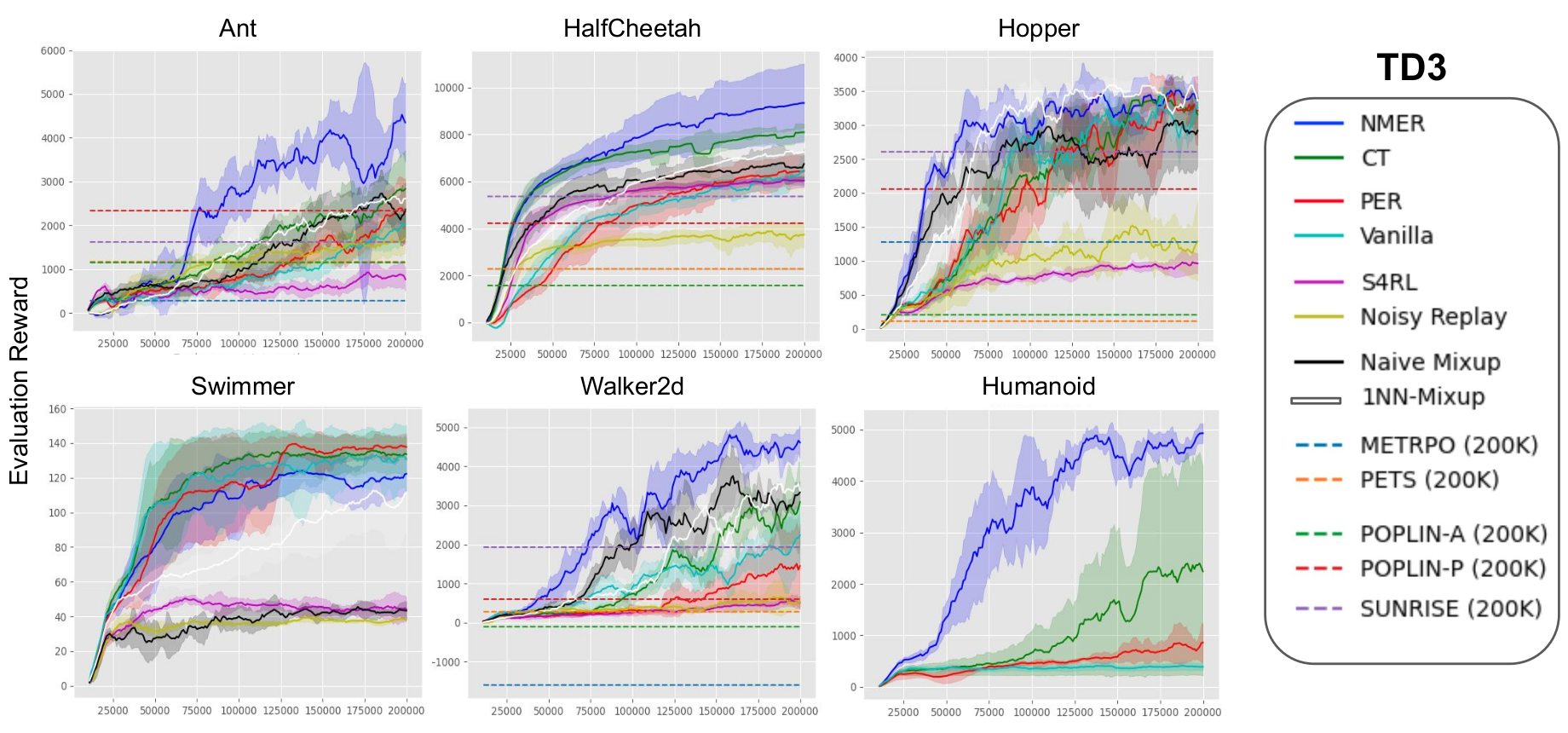}
    \caption{Learning curves for TD3 agents trained on NMER and baselines. Each replay buffer is run with four random seeds, and we plot mean performance with $\pm 1\sigma$ intervals.}
    \label{fig:td3_learning_curves}
\end{figure}

\begin{figure}[ht]
    \centering
    \includegraphics[width=1.00\textwidth]{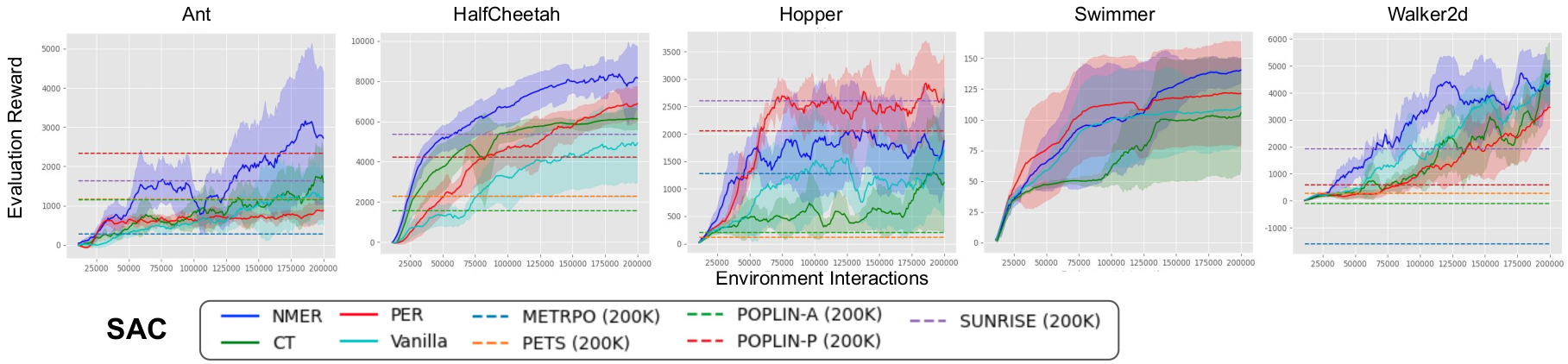}
    \caption{Learning curves for SAC agents trained on NMER and baselines. Each replay buffer is run with four random seeds, and we plot mean performance with $\pm 1\sigma$ intervals.}
    \label{fig:sac_learning_curves}
\end{figure}

Results for these evaluations are provided in Tables \ref{tab:results}, \ref{tab:sunrise_results}, and \ref{tab:compare_results}. Note the following abbreviations in Tables \ref{tab:results} and \ref{tab:compare_results}: (i) \texttt{U} = Uniform Replay, (ii) \texttt{PER} = Prioritized Experience Replay, (iii) \texttt{CT} = Continuous Transition, (iv) \texttt{NMER} = Neighborhood Mixup Experience Replay, (v) \texttt{1-NN Mixup} = NMER with one neighbor, (vi) \texttt{Mixup} = NMER with all neighbors (Naive Mixup), (vii) \texttt{S4RL} = S4RL, and (viii) $\mathcal{N}(0, \sigma^2)$ = Noisy Replay. Best results for each off-policy algorithm (\texttt{TD3}, \texttt{SAC}) are bolded. Detailed and additional results can be found in our \href{https://rmsander.github.io/projects/nmer_tech_report.pdf}{technical report}. Learning curves for \texttt{TD3} and \texttt{SAC} agents across all evaluated replay buffers are depicted in Figures \ref{fig:td3_learning_curves} and \ref{fig:sac_learning_curves} (\texttt{SUNRISE} and other baselines evaluated in \parencite{lee2020sunrise} are depicted with dashed horizontal lines indicating mean evaluation reward at 200K environment interactions). Each result is averaged over four runs.

The results of this continuous control evaluation study indicate that NMER frequently achieves comparatively better sample efficiency than the baseline replay buffers used in this study across SAC and TD3, as well as other baseline DRL algorithms evaluated in \parencite{lee2020sunrise}.
\begin{table}[ht]
\caption{Continuous control results from OpenAI Gym MuJoCo, 200K env. interactions.}
\label{tab:results}
\begin{center}
\small
\begin{tabular}{llllll}
\hline
\textbf{RL Agent} & \textbf{Environment} & \texttt{U} & \texttt{PER} & \texttt{CT} & \texttt{NMER} \\
\hline
\multirow{5}{*}{\texttt{TD3}} & 
\texttt{Ant} & 2005 $\pm$ 399 & 2317 $\pm$ 756 & 2834 $\pm$ 875 & \textbf{4347 $\pm$ 908}  \\
& \texttt{HalfCheetah} & 6467 $\pm$ 658 & 6447 $\pm$ 693 & 8097 $\pm$ 358 & \textbf{9340 $\pm$ 1678} \\
& \texttt{Hopper} & 3252 $\pm$ 157 & 3213 $\pm$ 511 & 3156 $\pm$ 351 & \textbf{3393 $\pm$ 220} \\
& \texttt{Swimmer} & 131 $\pm$ 20 & \textbf{138 $\pm$ 8} & 134 $\pm$ 10 & 122 $\pm$ 10 \\
& \texttt{Walker2d} & 2236 $\pm$ 686 & 1452 $\pm$ 1057 & 3087 $\pm$ 1058 & \textbf{4611 $\pm$ 441} \\
& \texttt{Humanoid} & 388 $\pm$ 66 & 860 $\pm$ 385 & 2242 $\pm$ 2027 & \textbf{4930 $\pm$ 190} \\
& $\Delta$ \texttt{NMER} (\%)& -37.5$\%$  & -36.8$\%$ & -22.1$\%$ & 0$\%$ \\
\hline
\multirow{5}{*}{\texttt{SAC}} & 
\texttt{Ant} & 1188 $\pm$ 692 & 800 $\pm$ 160 & 1594 $\pm$ 717 & \textbf{2721 $\pm$ 1685} \\
& \texttt{HalfCheetah} & 4918 $\pm$ 1928 & 6880 $\pm$ 886 & 6120 $\pm$ 525 & \textbf{8168 $\pm$ 1585} \\
& \texttt{Hopper} & 1692 $\pm$ 1160 & \textbf{2801 $\pm$ 827} & 1115 $\pm$ 897 & 1875 $\pm$ 900 \\
& \texttt{Swimmer} & 110 $\pm$ 29 & 121 $\pm$ 42 & 106 $\pm$ 46 & \textbf{140 $\pm$ 10} \\
& \texttt{Walker2d} & 4303 $\pm$ 636 & 3466 $\pm$ 784 & \textbf{4696 $\pm$ 1194} & 4429 $\pm$ 819 \\
& $\Delta$ \texttt{NMER} (\%)& -26.0$\%$  & -14.4$\%$ & -25.1$\%$ & 0$\%$ \\
\hline
\end{tabular}
\end{center}
\end{table}


\begin{table}[ht]
\caption{Baselines from SUNRISE \parencite{lee2020sunrise} vs TD3 + NMER, 200K env. interactions.}
\label{tab:sunrise_results}
\begin{center}
\small
\begin{tabular}{lllllll}
\hline
\textbf{Environment} & \texttt{METRPO} & \texttt{PETS} & \texttt{POPLIN-A} & \texttt{POPLIN-P} & \texttt{SUNRISE} & \texttt{NMER+TD3}\\
\hline
\texttt{Ant} & 282$\pm$18 & 1166$\pm$227 & 1148$\pm$438 & 2330$\pm$321 & 1627$\pm$293 & \textbf{4347$\pm$908}\\
\texttt{HalfCheetah} & 2284$\pm$900 & 2288$\pm$1019 & 1563$\pm$1137 & 4235$\pm$1133 & 5371$\pm$483 & \textbf{9340$\pm$1678} \\
\texttt{Hopper} & 1273$\pm$501 & 115$\pm$621 & 203$\pm$963 & 2055$\pm$614 & 2602$\pm$307 & \textbf{3393$\pm$220}\\
\texttt{Walker2d} & -1609$\pm$658 & 283$\pm$502 & -105$\pm$250 & 597$\pm$479 & 1926$\pm$695 & \textbf{4611$\pm$441}  \\
$\Delta$ \texttt{NMER} (\%)& -82.8$\%$  & -84.8$\%$ & -87.7$\%$ & -56.9$\%$ & -46.7$\%$ & 0$\%$\\
\hline
\end{tabular}
\end{center}
\end{table}
\begin{table}[ht]
\begin{center}
\caption{Additional baseline comparison study with TD3, 200K env. interactions.}
\label{tab:compare_results}
\small
\begin{tabular}{llllll}
\hline
\textbf{Environment} & \texttt{1NN-Mixup} & \texttt{Mixup} & \texttt{S4RL} & $\mathcal{N}(0, \sigma^2)$ & \texttt{NMER} \\
\hline
\texttt{Ant} & 	2651 $\pm$ 828 & 2361 $\pm$ 616	& 769 $\pm$ 270 & 1709 $\pm$ 350 & \textbf{4347 $\pm$ 908} \\
\texttt{HalfCheetah} & 7255 $\pm$ 1014 &  6743 $\pm$ 657 & 6032 $\pm$ 187 & 3733 $\pm$ 540 & \textbf{9340 $\pm$ 1678}\\
\texttt{Hopper} & 3360 $\pm$ 217 & 2917 $\pm$ 523 & 960 $\pm$ 151 & 1287 $\pm$ 593 & \textbf{3393 $\pm$ 220}\\
\texttt{Swimmer} & 111 $\pm$ 22 & 43 $\pm$ 5 & 44 $\pm$ 7 & 39 $\pm$ 1 &  \textbf{122 $\pm$ 10}\\
\texttt{Walker2d} &	3372 $\pm$ 833 & 3340 $\pm$ 293 & 514 $\pm$ 169 & 516 $\pm$ 157 & \textbf{4611 $\pm$ 441} \\
$\Delta$ \texttt{NMER} (\%) & -22.9\% & -36.0\% & -68.4\% & -67.9\% & 0\%  \\
\hline
\end{tabular}
\end{center}
\end{table}


\section{Discussion}
\label{sec:discussion}
These evaluation studies demonstrate that agents trained using NMER can learn robust policies using fewer environment interactions compared to agents trained using state-of-the-art replay buffers for continuous control. We consider the implications of NMER and its extensions to DRL.

\paragraph{Limitations.}\label{para:limitations} Despite the observed empirical success, NMER exhibits several limitations. NMER assumes the underlying transition dynamics are locally linear, which may be a naive approximation for continuous control tasks with nonlinear underlying dynamics. Additionally, although NMER's use of nearest neighbors serves as a viable heuristic to steer the replay buffer toward on-manifold interpolation, on-manifold interpolation is not analytically guaranteed. The sections below aim to put these limitations into context by suggesting viable generalizations and considerations for future work. Third, our experiments illustrate the limits of high replay (update-to-data) ratios, which could be explained by overestimation bias of the value function(s) \parencite{chen2020randomized, hiraoka2021dropout}.

\paragraph{Generalizing Synthetic Training.} As NMER uses convex interpolation to generate transitions for training, as more samples are added to the replay buffer, the interpolated transitions will become more accurate. Furthermore, NMER can be extended to modulate the ratio of real to synthetic training samples over time, enabling a variety of flexible training schemes for different replay buffer densities.

\paragraph{Generalizing neighborhoods.} NMER computes nearest neighbors using standardized Euclidean norms over concatenated state-action space, which allows for Mixup-based interpolation of transitions with proximal state-action vectors, regardless of the rewards and next states in these transitions. However, this notion of proximity between transitions can be generalized to any measure of distance in the transition space of a replay buffer. Generalizing this notion of proximity between stored environment interactions can be invoked via different distance metrics, e.g. Mahalanobis distance, as well as the use of composite product norms over different features in the transition space. For instance, the use of a composite product norm over states $\times$ actions results in nearest neighbors having similar (state, action) pairs. The efficacy of different proximity representations for neighborhood-based interpolated experience replay remains an open research question.

\paragraph{Generalizing interpolation.} NMER also invokes the implicit prior that linear combinations of states and actions result in the same linear combination of rewards and next states through the use of Mixup. However, under some dynamics regimes and continuous control environments, local linear interpolation via local Mixup may result in interpolated samples far from the underlying transition manifold. For interpolating on-manifold samples in locally nonlinear neighborhoods, off-policy DRL agents may benefit from the use of more sophisticated neighborhood-based interpolation mechanisms, such as Gaussian Process Regression \parencite{rasmussen2003gaussian, rasmussen2004gaussian} or Graph Neural Networks \parencite{zhou2020graph}.

\section{Conclusion}
We present Neighborhood Mixup Experience Replay (NMER), an experience replay buffer that improves the sample efficiency of off-policy DRL agents through synthetic sample generation. We empirically demonstrate that training agents on experiences generated via local Mixup in the transition space of a replay buffer facilitates learning robust policies using fewer environment interactions. NMER combines the benefits of learning from locally linear approximations of the underlying environment model with the sample efficiency benefits of learning from synthetic samples, thus expanding the possibilities for tractable DRL in real-world continuous control settings.

\acks{This research was supported by the Toyota Research Institute (TRI). This article solely reflects the opinions and conclusions of its authors and not TRI, Toyota, or any other entity. We thank TRI for their support. The authors thank the MIT SuperCloud and Lincoln Laboratory Supercomputing Center \parencite{reuther2018interactive} for providing HPC and consultation resources that have contributed to the research results reported within this publication.}

\printbibliography
\newpage
\appendix

\section{Off-Policy DRL algorithms}
In this section, we briefly discuss the two off-policy learning algorithms, Soft Actor-Critic (SAC) \parencite{pmlr-v80-haarnoja18b} and Twin Delayed Deep Deterministic Policy Gradients (TD3) \parencite{fujimoto2018addressing}.

\paragraph{Twin Delayed Deep Deterministic Policy Gradients} TD3 is a model-free, off-policy deep reinforcement learning algorithm. TD3 expands upon Deep Deterministic Policy Gradients (DDPG) \parencite{journals/corr/LillicrapHPHETS15} by employing a clipped double-Q trick to stabilize the actor's learning \parencite{fujimoto2018addressing}. The actor itself is parameterized as a neural mean function approximator $\mu_{\boldsymbol{\theta}}(s)$ that predicts optimal mean actions given a state $s$. We build on the TD3 implementation provided by RLlib \parencite{liang2018rllib}.

\paragraph{Soft Actor-Critic} SAC is also a model-free, off-policy deep reinforcement learning algorithm. It aims to maximize expected improvement on an entropy-regularized objective and trades off exploration with exploitation. Here, we optimize the coefficient $\alpha$ and leverage the clipped double-Q trick that TD3 makes use of to stabilize learning. We build on the SAC implementation provided by RLlib \parencite{liang2018rllib}.

\section{Nearest neighbor computation in NMER}
\label{app:nn_nmer}
\begin{figure}[H]
    \centering
    \includegraphics[width=0.5\textwidth]{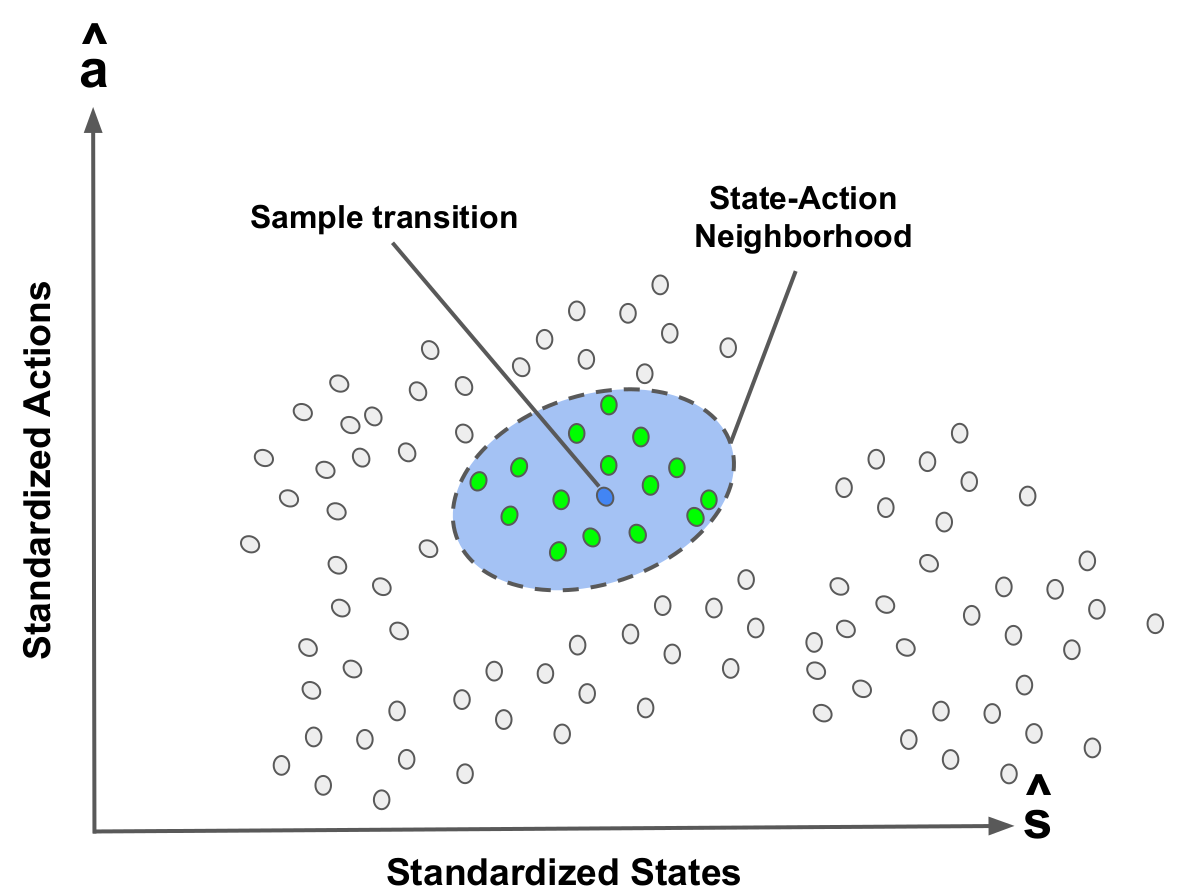}
    \caption{To compute nearest neighbors, NMER finds the closest transitions in the replay buffer to a given transition also sampled from the replay buffer in standardized state and action coordinates.}
    \label{fig:nearest_neighbors}
\end{figure}

In Neighborhood Mixup Experience Replay (NMER), nearest neighbors are computed using a concatenated state-action vector norm in tandem with Euclidean distance search over the Z-score standardized states and actions. Figure \ref{fig:nearest_neighbors} illustrates this.

FAISS \parencite{JDH17} is used to compute nearest neighbors using a C++ backend in tandem with GPU-based acceleration with CUDA \parencite{cuda}. Relative runtimes of NMER are given as averaged over four seeds in the \texttt{Humanoid-v2} environment using our 200K environment interaction benchmark. As can be seen, NMER introduces minimal computational overhead, even for high-dimensional state and action spaces such as \texttt{Humanoid-v2}.

\section{Baselines}
For our comparative evaluation studies, we implemented and evaluated several baseline replay buffers used in state-of-the-art off-policy experience replay.

\paragraph{Uniform Replay (\texttt{U})} Uniform Replay \parencite{mnih2013playing} is a circular replay buffer that selects stored transitions uniformly at random, in an independent and identically-distributed fashion. If the sample limit of this replay buffer is reached, samples are replaced in a first-in, first-out (FIFO) fashion.

\paragraph{Prioritized Experience Replay (\texttt{PER})} PER \parencite{schaul2016prioritized} extends Uniform Replay by selecting samples for training the DRL agent in a prioritized fashion, rather than uniformly at random. PER prioritizes these samples according to their estimated TD-error, which, intuitively, is a heuristic estimate for the ``surprise'', and therefore learnability, a given transition induces on the DRL agent. For these comparative evaluation experiments, we make use of the stochastic prioritization variant of PER, in which samples are selected stochastically with sampling weights proportional to their priority. We make use of the PER implementation provided by \parencite{liang2018rllib}.

\paragraph{Continuous Transition (\texttt{CT})} CT \parencite{lin2020continuous} is a replay buffer interpolation and data augmentation module designed for continuous state and action spaces. Specifically, it interpolates temporally-adjacent transitions, on the condition that they fall in the same episode. CT utilizes this temporal proximity as a heuristic, in tandem with a Mixup temperature ($\alpha$) discriminator to ensure that interpolated transitions remain approximately on the transition manifold \parencite{lin2020continuous}. We implement our own variant of CT, as described in the experimental section of the manuscript.

\section{Additional Baselines}
In addition to the baselines considered above, we also considered baselines to intuitively and rigorously demonstrate the strong empirical performance of NMER. These baselines include:

\begin{itemize}
    \item \textbf{Noisy Replay} ($\mathcal{N}(0, \sigma^2)$): Perturbing the original transitions used to train the reinforcement learning agents with zero-mean, i.i.d. Gaussian noise. Below,  is the standard deviation of the zero-mean i.i.d. Gaussian noise.
    \item \textbf{Naive Mixup} (\texttt{Mixup}): Standard Mixup - this is the limiting case where neighborhood size.
    \item \textbf{S4RL} (\texttt{S4RL}): The S4RL observation/next observation Mixup technique \parencite{sinha2021srl}.
    \item \textbf{1-Nearest Neighbor Mixup} (\texttt{1NN-Mixup}): The other extreme neighborhood case. This baseline is paired with baseline 3 to better empirically quantify the effect of neighborhood size on agent performance.
\end{itemize}

\section{Other Ablative Factors Considered}
In addition to running ablations over the replay ratio variable, we also consider the following ablation factors for future work in NMER:

\begin{enumerate}
    \item \textbf{Neighborhood Size $k$}: Mixup and neighborhood sampling stand in somewhat stark contrast to one another - while Mixup encourages generalization by interpolating any pair of samples in the dataset \parencite{lin2020continuous}, invoking a particularly strong inductive bias, neighborhood sampling reduces the degree of this inductive bias by restricting this bias to proximal neighborhoods of the inputs on the transition manifold. Systematically evaluating this trade-off can help to yield optimal $k$ values for continuous control tasks and baseline RL algorithms (e.g. SAC, TD3).
    \item \textbf{Fraction of Samples Interpolated}: Another ablative factor that can be used to measure the efficacy of NMER relative to the underlying baseline RL algorithms is the fraction of samples to interpolate in a mini-batch. An ablation study over this interpolation fraction can indicate whether a combination of observed and interpolated transitions results in better performance compared to entirely interpolated or entirely observed transitions.
\end{enumerate}

\section{Reinforcement Learning Agent Hyperparameters}
The hyperparameters in this section detail the hyperparameterization for the deep reinforcement learning agents applied for NMER. Namely, these hyperparameterizations are for Soft Actor-Critic (SAC) \parencite{pmlr-v80-haarnoja18b} and Twin Delayed Deep Deterministic Policy Gradients (TD3) \parencite{fujimoto2018addressing}. Environment-specific parameters are provided in the following section.
\subsection{Soft Actor-Critic (SAC) Parameters}
Hyperparameters used for running comparative and ablation evaluation experiments on SAC \parencite{pmlr-v80-haarnoja18b} are provided in Table \ref{tab:sac_hyper}.
\begin{table}[ht]
    \centering
    \caption{Table of hyperparameters for Soft Actor-Critic (SAC) across the different environments on which these replay buffers were evaluated. These hyperparameters are adapted from \parencite{liang2018rllib}.}
    \begin{tabular}{ll}
         \hline
         \textbf{Hyperparameter} & \textbf{Value} \\
         \hline
         Actor Learning Rate & 0.003  \\
         \hline
         Critic Learning Rate & 0.003  \\
         \hline
         Alpha Learning Rate & 0.003  \\
         \hline
         Polyak (target network update coefficient) $\tau$ & 0.005 \\
         \hline
         Target Network Update Interval (Gradient Steps) & 1 \\
         \hline
         Entropy Target & `auto' ($\dim(\mathcal{A})$)\\
         \hline
         Initial Entropy Parameter ($\alpha$) & 1.0 \\
         \hline
         Twin-Q & True \\
         \hline
         Normalize Actions & True \\
         \hline
         N-step & 1 \\
         \hline
         Gamma ($\gamma$) & 0.99 \\
         \hline
         Policy Parameterization & Squashed Gaussian \\
         \hline
         Clip Actions & False \\
         \hline
         Critic Hidden Units & [256, 256] \\
         \hline
         Critic activation function & \texttt{ReLU} \\
         \hline
         Actor Hidden Units & [256, 256] \\
         \hline
         Actor activation function & \texttt{ReLU} \\
         \hline
         Train batch size & 256 \\
         \hline
         Replay Buffer Size & 1000000 \\
         \hline
    \end{tabular}
    \label{tab:sac_hyper}
\end{table}
\subsubsection{SAC L2 regularization}
In addition to the SAC hyperparameters given in Table \ref{tab:sac_hyper}, we empirically observe that all replay buffer variants, including Vanilla Replay, result in actor network divergence when combined with SAC and high replay ratios in many of the continuous control environments considered. To mitigate this divergence, we apply a small L2 regularization penalty to the actor network of the SAC agent. The ablation studies above detail this regularization value as `L' for each replay buffer variant in each continuous control environment.

\subsection{Twin Delayed Deep Deterministic Policy Gradients (TD3)}
Hyperparameters used for running comparative and ablation evaluation experiments on TD3 \parencite{fujimoto2018addressing} are provided in Table \ref{tab:td3_hyper}.
\begin{table}[ht]
    \caption{Table of hyperparameters for Twin Delayed Deep Deterministic Policy Gradients (TD3) across the different environments on which these replay buffers were evaluated. These hyperparameters are adapted from \parencite{liang2018rllib}.}
    \centering
    \begin{tabular}{ll}
         \hline
         \textbf{Hyperparameter} & \textbf{Value} \\
         \hline
         Actor Learning Rate & 0.0005 \\
         \hline
         Critic Learning Rate & 0.0005  \\
         \hline
         Polyak (target network update coefficient) $\tau$ & 0.005 \\
         \hline
         Target Network Update Interval (Gradient Steps) & 1 \\
         \hline
         Twin-Q & True \\
         \hline
         Normalize Actions & True \\
         \hline
         Policy Parameterization & Squashed Gaussian \\
         \hline
         Policy Delay & 2 \\
         \hline
         Smooth Target Policy & True \\
         \hline
         Target Noise & 0.2 \\
         \hline
         Target Noise Clip & 0.5 \\
         \hline
         Exploration Noise Type & $\mathcal{N}(0, 0.1)$ \\
         \hline
         Random Steps & 10000 \\
         \hline 
         N-step & 1 \\
         \hline
         Gamma ($\gamma$) & 0.99 \\
         \hline
         Clip Actions & False \\
         \hline
         Critic Hidden Units & [400, 300] \\
         \hline
         Actor Hidden Units & [400, 300] \\
         \hline
         Train batch size & 100 \\
         \hline
         L2 Regularization & 0 \\
         \hline
         Replay Buffer Size & 1000000 \\
         \hline
    \end{tabular}
    \label{tab:td3_hyper}
\end{table}
\subsection{Prioritized Experience Replay Hyperparameters}
Table \ref{tab:per_hyper} details the hyperparameter configurations for the stochastic prioritization variant of Prioritized Experience Replay (PER) \parencite{schaul2016prioritized} used for these replay buffers and variants. PER hyperparameters were kept constant for all environments and algorithms. We build on the PER implementation provided by RLlib \parencite{liang2018rllib}.
\begin{table}[ht]
    \centering
    \begin{tabular}{ll}
         \hline
         \textbf{Hyperparameter} & \textbf{Value} \\
         \hline
         $\alpha$ & 0.6 \\
         \hline
         Initial $\beta$ & 0.4 \\
         \hline
         Final $\beta$ & 0.4 \\
         \hline
         $\beta$ Annealing Time Steps & 20000 \\
         \hline
         $\epsilon$ & 0.000001 \\
         \hline
    \end{tabular}
    \caption{Table of Prioritized Experience Replay (PER) \parencite{schaul2016prioritized} hyperparameters used for all comparative and evaluation studies with the PER baseline. These hyperparameters are adapted from \parencite{liang2018rllib}.}
    \label{tab:per_hyper}
\end{table}

\section{Environment details}
For our comparative evaluation and ablation studies, we use the OpenAI MuJoCo continuous control task suite, as well as the OpenAI Classic Control suite. Dimensions of the continuous state and action spaces for each of these environments, as well as whether termination signals are automatically applied at the end of an epiosde, can be found in Table \ref{tab:state_action_dim}.

\begin{table}[ht]
    \centering
    \caption{State and action space dimensions for the continuous control environments we ran comparative evaluation and ablation studies on with replay buffers.}
    \begin{tabular}{cccc}
         \hline
         \textbf{Environment} & Number of States & Number of Actions & No Done At End? \\
         \hline
         \texttt{Ant-v2} & 111 & 8 & False\\
         \hline
         \texttt{HalfCheetah-v2} & 17 & 6 & True\\
         \hline
         \texttt{Hopper-v2} & 8 & 3 & False\\
         \hline
         \texttt{Swimmer-v2} & 8 & 2 & True\\
         \hline
         \texttt{Walker2d-v2} & 17 & 6 & False\\
         \hline
         \texttt{Humanoid-v2} & 376 & 17 & False\\
         \hline
    \end{tabular}
    \label{tab:state_action_dim}
\end{table}

\subsection{Determining appropriate episode termination signals}
For these continuous control tasks, it is important to determine appropriate episode termination signals provided to the agent in order for them to properly learn which steps mark the end of an episode. Since many of these continuous control tasks are infinite-horizon in nature, simply setting the final step of an episode to have a termination signal may inadvertently train the agent to avoid this state-action pair in the future, since it is treated as the last step regardless of whether early episode termination conditions (e.g. the \texttt{Walker2d-v2} agent falling over) actually apply. On the contrary, it is also important that agents are provided with information related to termination signals when true termination conditions actually apply.

Therefore, in general, we turn termination signals off for these environments. For environments in which early termination signals can apply (e.g. the \texttt{Walker2d-v2} agent falling over), we allow for termination signals, but only apply them if the termination signal occurs before the horizon. If the termination signal occurs at the horizon, we do not use a termination signal, since this implies this step is the last in the horizon. Table \ref{tab:state_action_dim} shows whether a termination signal is applied automatically on the final step of the environment. This additional conditional termination logic is applied on top of environments in which `No Done At End?' is set to `False'.

Crucially, this logic is applied to all variants tested for our comparative evaluation and ablation studies in order to ensure fairness to the different replay buffer variants being evaluated.

\section{Detailed Ablation Studies, OpenAI Gym}
Ablation studies were to determine the optimal replay ratio for each replay buffer variant. Since this optimal replay ratio differs for each replay variant as well as for each environment, it was important to run ablations over several replay ratios, namely 1, 5, and 20. The graphs below provide insight as to how the optimal replay buffer was chosen. These graphs are divided according to the underlying deep reinforcement learning algorithm run. 

Additionally, for Soft Actor-Critic experiments, ablation studies were performed to determine approximately minimal actor network L2 regularization needed in order to ensure policy stability. We empirically observe actor network divergence, particularly for higher replay ratios, in the absence of actor network regularization for SAC. In nearly all environments, the L2 regularization coefficient is set to be smaller for the baseline replay buffers than for NMER, in order to ensure that NMER was not gaining an unfair advantage through having less constraints on its neural policy.

Summarized, these ablation factors are:

\begin{enumerate}
    \item \textbf{Replay ratio (RR)}: The number of gradient training steps for every environment interaction the agent has with its environment. The replay ratios tested in these experiments were 1, 5, and 20. These replay ratio values are denoted by `RR'.
    \item \textbf{(SAC-only) L2 Actor Network Regularization (L)}: The L2 regularization coefficient used to ensure actor network stability for the Soft Actor-Critic studies. These values are denoted by `L'. Additionally, for the \texttt{Hopper-v2} experiments, we consider a gradient clipping value of 40 (SAC-only).
\end{enumerate}

These values are given in each of the graphs and tables below.

\subsection{TD3 Ablation Studies}
For each environment and replay buffer variant (\texttt{U}, \texttt{PER}, \texttt{CT}, and \texttt{NMER}), we run ablation studies in which we vary the replay ratio to determine the optimal replay ratio for each variant. These ablation studies are conducted to ensure that we compare NMER against competitive baselines. Empirically, these ablation studies are motivated by observing that for one subset of the following continuous control tasks, our baseline replay buffers perform better with lower replay ratios, and for another subset of these continuous control tasks, these same baseline replay buffers perform better with higher replay ratios.

Results from each (environment, replay buffer) variant are given in the following tables and figures. The best ablation result for each replay buffer variant is shown in bold.

\begin{table}[ht]
\begin{center}
\caption{Evaluation reward for TD3 and \texttt{Ant-v2} outfitted with replay buffers with varying replay ratios.}
\begin{tabular}{lc}
\hline
\textbf{Replay Buffer} & \textbf{Reward (200K steps)} \\
\hline\hline
\texttt{U, RR=1} & \textbf{2005 $\pm$ 399} \\
\texttt{U, RR=5} & -12 $\pm$ 3 \\
\texttt{U, RR=20} & -50 $\pm$ 5 \\
\texttt{PER, RR=1} \parencite{schaul2016prioritized} & \textbf{2317 $\pm$ 756} \\
\texttt{PER, RR=5} \parencite{schaul2016prioritized} & -19 $\pm$ 4 \\
\texttt{PER, RR=20} \parencite{schaul2016prioritized} & -50 $\pm$ 9 \\
\texttt{CT, RR=1} \parencite{lin2020continuous} & \textbf{2834 $\pm$ 875} \\
\texttt{CT, RR=5} \parencite{lin2020continuous} & 1986 $\pm$ 1750 \\
\texttt{CT, RR=20} \parencite{lin2020continuous} & 3 $\pm$ 6 \\
\textbf{\texttt{NMER, K=25, RR=20}} & \textbf{4347 $\pm$ 908} \\
\hline
\end{tabular}
\end{center}
\label{tab:ant_td3}
\end{table}
\begin{table}[ht]
\begin{center}
\caption{Evaluation reward for TD3 and \texttt{HalfCheetah-v2} outfitted with replay buffers with varying replay ratios.}
\begin{tabular}{lc}
\hline
\textbf{Replay Buffer} & \textbf{Reward (200K steps)} \\
\hline\hline
\texttt{U, RR=1} & \textbf{6467 $\pm$ 658} \\
\texttt{U, RR=5} & 6295 $\pm$ 560 \\
\texttt{U, RR=20} & 1059 $\pm$ 574 \\
\texttt{PER, RR=1} \parencite{schaul2016prioritized} & 6388 $\pm$ 474 \\
\texttt{PER, RR=5} \parencite{schaul2016prioritized} & \textbf{6447 $\pm$ 693} \\
\texttt{PER, RR=20} \parencite{schaul2016prioritized} & 2542 $\pm$ 1078 \\
\texttt{CT, RR=1} \parencite{lin2020continuous} & 6821 $\pm$ 745 \\
\texttt{CT, RR=5} \parencite{lin2020continuous} & 6831 $\pm$ 997 \\
\texttt{CT, RR=20} \parencite{lin2020continuous} & \textbf{8097 $\pm$ 358} \\
\textbf{\texttt{NMER, K=10, RR=20}} & \textbf{9327 $\pm$ 1099} \\
\hline
\end{tabular}
\end{center}
\label{tab:td3_halfcheetah}
\end{table}
\begin{table}[ht]
\begin{center}
\caption{Evaluation reward for TD3 and \texttt{Hopper-v2} outfitted with replay buffers with varying replay ratios.}
\begin{tabular}{lc}
\hline
\textbf{Replay Buffer} & \textbf{Reward (200K steps)} \\
\hline\hline
\texttt{U, RR=1} & 2371 $\pm$ 593 \\
\texttt{U, RR=5} & 2758 $\pm$ 607 \\
\texttt{U, RR=20} & \textbf{3252 $\pm$ 157} \\
\texttt{PER, RR=1} \parencite{schaul2016prioritized} & 2860 $\pm$ 387 \\
\texttt{PER, RR=5} \parencite{schaul2016prioritized} & \textbf{3213 $\pm$ 511} \\
\texttt{PER, RR=20} \parencite{schaul2016prioritized} & 2190 $\pm$ 530 \\
\texttt{CT, RR=1} \parencite{lin2020continuous} & 1891 $\pm$ 825 \\
\texttt{CT, RR=5} \parencite{lin2020continuous} & \textbf{3156 $\pm$ 351} \\
\texttt{CT, RR=20} \parencite{lin2020continuous} & 3080 $\pm$ 437 \\
\textbf{\texttt{NMER, K=10, RR=20}} & \textbf{3411 $\pm$ 340}\\
\hline
\end{tabular}
\end{center}
\label{tab:td3_hopper}
\end{table}
\begin{table}[ht]
\begin{center}
\caption{Evaluation reward for TD3 and \texttt{Swimmer-v2} outfitted with replay buffers with varying replay ratios.}
\begin{tabular}{lc}
\hline
\textbf{Replay Buffer} & \textbf{Reward (200K steps)} \\
\hline\hline
\texttt{U, RR=1} & 56 $\pm$ 8 \\
\texttt{U, RR=5} & 92 $\pm$ 25\\
\texttt{U, RR=20} & \textbf{131 $\pm$ 20} \\
\texttt{PER, RR=1} \parencite{schaul2016prioritized} & 64 $\pm$ 12 \\
\texttt{PER, RR=5} \parencite{schaul2016prioritized} & 107 $\pm$ 31 \\
\texttt{PER, RR=20} \parencite{schaul2016prioritized} & \textbf{138 $\pm$ 8} \\
\texttt{CT, RR=1} \parencite{lin2020continuous} & 60 $\pm$ 10 \\
\texttt{CT, RR=5} \parencite{lin2020continuous} & 118 $\pm$ 20 \\
\texttt{CT, RR=20} \parencite{lin2020continuous} & \textbf{134 $\pm$ 10} \\
\textbf{\texttt{NMER, K=10, RR=20}} & \textbf{131 $\pm$ 20} \\
\hline
\end{tabular}
\end{center}
\label{tab:hopper_td3}
\end{table}
\begin{table}[ht]
\begin{center}
\caption{Evaluation reward for TD3 and \texttt{Walker2d-v2} outfitted with replay buffers with varying replay ratios.}
\begin{tabular}{lc}
\hline
\textbf{Replay Buffer} & \textbf{Reward (200K steps)} \\
\hline\hline
\texttt{U, RR=1} & \textbf{2236 $\pm$ 686} \\
\texttt{U, RR=5} & 1833 $\pm$ 934 \\
\texttt{U, RR=20} & 682 $\pm$ 806 \\
\texttt{PER, RR=1} \parencite{schaul2016prioritized} & 1006 $\pm$ 332 \\
\texttt{PER, RR=5} \parencite{schaul2016prioritized} & \textbf{1452 $\pm$ 1057} \\
\texttt{PER, RR=20} \parencite{schaul2016prioritized} & 476 $\pm$ 646 \\
\texttt{CT, RR=1} \parencite{lin2020continuous} & 2198 $\pm$ 770 \\
\texttt{CT, RR=5} \parencite{lin2020continuous} & \textbf{3087 $\pm$ 1058} \\
\texttt{CT, RR=20} \parencite{lin2020continuous} & 3079 $\pm$ 1331 \\
\textbf{\texttt{NMER, K=10, RR=20}} & \textbf{3960 $\pm$ 777} \\
\hline
\end{tabular}
\end{center}
\label{tab:walker_td3}
\end{table}
\begin{table}[ht]
\begin{center}
\caption{Evaluation reward for TD3 and \texttt{Humanoid-v2} outfitted with replay buffers with varying replay ratios.}
\begin{tabular}{lc}
\hline
\textbf{Replay Buffer} & \textbf{Reward (200K steps)} \\
\hline\hline
\texttt{U, RR=1} & 293 $\pm$ 0 \\
\texttt{U, RR=5} & \textbf{388 $\pm$ 66} \\
\texttt{U, RR=20} & 385 $\pm$ 0 \\
\texttt{PER, RR=1} \parencite{schaul2016prioritized} & \textbf{860 $\pm$ 385} \\
\texttt{PER, RR=5} \parencite{schaul2016prioritized} & 476 $\pm$ 18\\
\texttt{PER, RR=20} \parencite{schaul2016prioritized} & 273 $\pm$ 123 \\
\texttt{CT, RR=1} \parencite{lin2020continuous} & 2122 $\pm$ 372\\
\texttt{CT, RR=5} \parencite{lin2020continuous} & \textbf{2242 $\pm$ 2027}\\
\texttt{CT, RR=20} \parencite{lin2020continuous} & 823 $\pm$ 0\\
\textbf{\texttt{NMER, K=10, RR=20}} & \textbf{4791 $\pm$ 271} \\
\hline
\end{tabular}
\end{center}
\label{tab:humanoid_td3}
\end{table}

\subsection{SAC Ablation Studies}

\begin{table}[ht]
\begin{center}
\caption{Evaluation reward for SAC and \texttt{Ant-v2} outfitted with replay buffers with varying replay ratios.}
\begin{tabular}{lc}
\hline
\textbf{Replay Buffer} & \textbf{Reward (200K steps)} \\
\hline\hline
\texttt{U, RR=1, L=1e-3} & 975 $\pm$ 63 \\
\texttt{U, RR=1, L=5e-4} & 1111 $\pm$ 235 \\
\texttt{U, RR=5, L=1e-3} & \textbf{1188 $\pm$ 692} \\
\texttt{U, RR=5, L=5e-4} & 822 $\pm$ 120\\
\texttt{U, RR=20, L=1e-3} & 610 $\pm$ 141\\
\texttt{PER, RR=1, L=1e-3} \parencite{schaul2016prioritized} & \textbf{878 $\pm$ 239} \\
\texttt{PER, RR=5, L=1e-3} \parencite{schaul2016prioritized} & 800 $\pm$ 160 \\
\texttt{PER, RR=1, L=5e-4} \parencite{schaul2016prioritized} & 655 $\pm$ 257 \\
\texttt{PER, RR=5, L=5e-4} \parencite{schaul2016prioritized} & 581 $\pm$ 184 \\
\texttt{CT, RR=1, L=1e-3} \parencite{lin2020continuous} & 1395 $\pm$ 481 \\
\texttt{CT, RR=5, L=1e-3} \parencite{lin2020continuous} & \textbf{1594 $\pm$ 717} \\
\texttt{CT, RR=20, L=1e-3} \parencite{lin2020continuous} & 1166 $\pm$ 388 \\
\texttt{CT, RR=1, L=5e-4} \parencite{lin2020continuous} & 1181 $\pm$ 547 \\
\texttt{CT, RR=5, L=5e-4} \parencite{lin2020continuous} & 1098 $\pm$ 472 \\
\texttt{NMER, K=25, RR=20, L=2e-3} & \textbf{2723 $\pm$ 1685} \\
\texttt{NMER, K=10, RR=20, L=5e-4} & 2507 $\pm$ 962 \\
\texttt{NMER, K=10, RR=5, L=5e-4} & 1342 $\pm$ 773 \\

\hline
\end{tabular}
\end{center}
\label{tab:ant_results}
\end{table}
\begin{table}[ht]
\begin{center}
\caption{Evaluation reward for SAC and \texttt{HalfCheetah-v2} outfitted with replay buffers with varying replay ratios.}
\begin{tabular}{lc}
\hline
\textbf{Replay Buffer} & \textbf{Reward (200K steps)} \\
\hline\hline
\texttt{U, RR=1, L=1e-3} & 3424 $\pm$ 2093 \\
\texttt{U, RR=5, L=1e-3} & \textbf{4918 $\pm$ 1928} \\
\texttt{U, RR=20, L=5e-4} & 4297 $\pm$ 1700 \\
\texttt{PER, RR=1, L=1e-3} \parencite{schaul2016prioritized} & 4665 $\pm$ 1926 \\
\texttt{PER, RR=5, L=1e-3} \parencite{schaul2016prioritized} & 6518 $\pm$ 1117 \\
\texttt{PER, RR=20, L=1e-3} \parencite{schaul2016prioritized} & 6444 $\pm$ 364 \\
\texttt{PER, RR=1, L=5e-4} \parencite{schaul2016prioritized} & 5002 $\pm$ 2093 \\
\texttt{PER, RR=5, L=5e-4} \parencite{schaul2016prioritized} & \textbf{6880 $\pm$ 886} \\
\texttt{PER, RR=20, L=5e-4} \parencite{schaul2016prioritized} & 5358 $\pm$ 415 \\
\texttt{PER, RR=20, L=1e-4} \parencite{schaul2016prioritized} & 6134 $\pm$ 891 \\
\texttt{CT, RR=1, L=1e-3} \parencite{lin2020continuous} & 5102 $\pm$ 592 \\
\texttt{CT, RR=5, L=1e-3} \parencite{lin2020continuous} & 4271 $\pm$ 1934 \\
\texttt{CT, RR=20, L=1e-3} \parencite{lin2020continuous} & \textbf{5263 $\pm$ 2146} \\
\texttt{CT, RR=1, L=5e-4} \parencite{lin2020continuous} & 4108 $\pm$ 1523 \\
\texttt{CT, RR=5, L=5e-4} \parencite{lin2020continuous} & 5121 $\pm$ 738 \\
\texttt{CT, RR=20, L=5e-4} \parencite{lin2020continuous} & 4894 $\pm$ 2002 \\
\texttt{NMER, K=10, RR=20, L=1e-3} & \textbf{8168 $\pm$ 1585} \\
\texttt{NMER, K=25, RR=20, L=1e-3} & 5812 $\pm$ 2619 \\
\hline
\end{tabular}
\end{center}
\label{tab:halfcheetah_results}
\end{table}
\begin{table}[ht]
\begin{center}
\caption{Evaluation reward for SAC and \texttt{Hopper-v2} outfitted with replay buffers with varying replay ratios.}
\begin{tabular}{lc}
\hline
\textbf{Replay Buffer} & \textbf{Reward (200K steps)} \\
\hline\hline
\texttt{U, RR=1, L=1e-2} & 427 $\pm$ 303 \\
\texttt{U, RR=5, L=1e-2} & 1333 $\pm$ 1006 \\
\texttt{U, RR=20, L=1e-2} & 763 $\pm$ 795 \\
\texttt{U, RR=5, L=1e-3, G=40} & \textbf{1692 $\pm$ 1160} \\
\texttt{PER, RR=1, L=1e-2} \parencite{schaul2016prioritized} & 2472 $\pm$ 444 \\
\texttt{PER, RR=5, L=1e-2} \parencite{schaul2016prioritized} & \textbf{2630 $\pm$ 843} \\
\texttt{PER, RR=5, L=1e-3} \parencite{schaul2016prioritized} & 2296 $\pm$ 946 \\
\texttt{PER, RR=5, L=5e-3, G=40} \parencite{schaul2016prioritized} & 2449 $\pm$ 452 \\
\texttt{CT, RR=1, L=1e-2} \parencite{lin2020continuous} & 412 $\pm$ 164 \\
\texttt{CT, RR=5, L=1e-2} \parencite{lin2020continuous} & 937 $\pm$ 592 \\
\texttt{CT, RR=20, L=5e-3, G=40} \parencite{lin2020continuous} & \textbf{1115 $\pm$ 897} \\
\texttt{NMER, K=10, RR=20, L=2e-2} & 1343 $\pm$ 424 \\
\texttt{NMER, K=10, RR=20, L=1e-2, G=40} & \textbf{1875 $\pm$ 900} \\

\hline
\end{tabular}
\end{center}
\label{tab:hopper_results}
\end{table}
\begin{table}[ht]
\begin{center}
\caption{Evaluation reward for SAC and \texttt{Swimmer-v2} outfitted with replay buffers with varying replay ratios.}
\begin{tabular}{lc}
\hline
\textbf{Replay Buffer} & \textbf{Reward (200K steps)} \\
\hline\hline
\texttt{U, RR=1, L=1e-4} & 51 $\pm$ 7 \\
\texttt{U, RR=5, L=1e-4} & 79 $\pm$ 25 \\
\texttt{U, RR=20, L=1e-4} & \textbf{111 $\pm$ 29} \\
\texttt{PER, RR=1, L=1e-4} \parencite{schaul2016prioritized} & 48 $\pm$ 1 \\
\texttt{PER, RR=5, L=1e-4} \parencite{schaul2016prioritized} & 114 $\pm$ 31 \\
\texttt{PER, RR=20, L=1e-4} \parencite{schaul2016prioritized} & \textbf{121 $\pm$ 42} \\
\texttt{CT, RR=1, L=1e-4} \parencite{lin2020continuous} & 51 $\pm$ 9 \\
\texttt{CT, RR=5, L=1e-4} \parencite{lin2020continuous} & 74 $\pm$ 20 \\
\texttt{CT, RR=20, L=5e-4} \parencite{lin2020continuous} & \textbf{106 $\pm$ 46} \\
\texttt{NMER, K=25, RR=20, L=1e-4} & \textbf{140 $\pm$ 10} \\
\texttt{NMER, K=10, RR=20, L=2e-4} & 95 $\pm$ 40 \\
\hline
\end{tabular}
\end{center}
\label{tab:swimmer_results}
\end{table}
\begin{table}[ht]
\begin{center}
\caption{Evaluation reward for SAC and \texttt{Walker2d-v2} outfitted with replay buffers with varying replay ratios.}
\begin{tabular}{lc}
\hline
\textbf{Replay Buffer} & \textbf{Reward (200K steps)} \\
\hline\hline
\texttt{U, RR=1, L=1e-2} & 1481 $\pm$ 546 \\
\texttt{U, RR=5, L=1e-2} & \textbf{4303 $\pm$ 636} \\
\texttt{U, RR=20, L=1e-2} & 3295 $\pm$ 901 \\
\texttt{PER, RR=1, L=1e-2} \parencite{schaul2016prioritized} & \textbf{3466 $\pm$ 784} \\
\texttt{PER, RR=5, L=1e-2} \parencite{schaul2016prioritized} & 3376 $\pm$ 611 \\
\texttt{PER, RR=20, L=1e-2} \parencite{schaul2016prioritized} & 2513 $\pm$ 728 \\
\texttt{CT, RR=1, L=1e-2} \parencite{lin2020continuous} & 1982 $\pm$ 920 \\
\texttt{CT, RR=5, L=1e-2} \parencite{lin2020continuous} & 3319 $\pm$ 898 \\
\texttt{CT, RR=20, L=1e-2} \parencite{lin2020continuous} & \textbf{4696 $\pm$ 1194} \\
\texttt{NMER, K=10, RR=20} & 3282 $\pm$ 813 \\
\texttt{NMER, K=25, RR=20} & \textbf{4429 $\pm$ 819} \\
\hline
\end{tabular}
\end{center}
\label{tab:walker2d_results}
\end{table}
\end{document}